# Bin Packing Optimization via Deep Reinforcement Learning

Baoying Wang, Huixu Dong, *Member, IEEE*

*Abstract*—The Bin Packing Problem (BPP) has attracted enthusiastic research interest recently, owing to widespread applications in logistics and warehousing environments. It is truly essential to optimize the bin packing to enable more objects to be packed into boxes. Object packing order and placement strategy are the two crucial optimization objectives of the BPP. However, existing optimization methods for BPP, such as the genetic algorithm (GA), emerge as the main issues in highly computational cost and relatively low accuracy, making it difficult to implement in realistic scenarios. To well relieve the research gaps, we present a novel optimization methodology of two-dimensional (2D)-BPP and three-dimensional (3D)-BPP for objects with regular shapes via deep reinforcement learning (DRL), maximizing the space utilization and minimizing the usage number of boxes. First, an end-to-end DRL neural network constructed by a modified Pointer Network consisting of an encoder, a decoder and an attention module is proposed to achieve the optimal object packing order. Second, conforming to the top-down operation mode, the placement strategy based on a height map is used to arrange the ordered objects in the boxes, preventing the objects from colliding with boxes and other objects in boxes. Third, the reward and loss functions are defined as the indicators of the compactness, pyramid, and usage number of boxes to conduct the training of the DRL neural network based on an on-policy actor-critic framework. Finally, a series of experiments are implemented to compare our method with conventional packing methods, from which we conclude that our method outperforms these packing methods in both packing accuracy and efficiency.

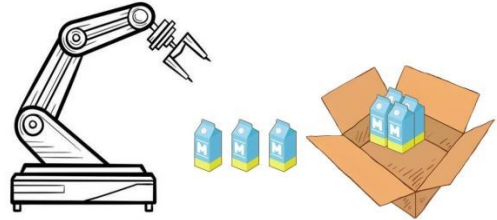

Figure. 1. The BPP is widely useful in realistic application scenarios, such as logistics and warehousing.

## I. INTRODUCTION

Known to be a classical strongly non-deterministic polynomial-time (NP) hard combinatorial optimization problem with high complexity, the Bin Packing Problem (BPP) has attracted considerable attention in realistic application scenarios, such as logistics and warehousing environments due to the rapid development of the industrial automation and the labor shortage [1, 2, 3] (see Fig.1). The key to optimizing the BPP lies in generating the object packing order and determining the object placement strategy. To maximize the space utilization and minimize the usage number of boxes, conventional packing methods [1, 4, 5] take advantage of the meta-heuristics to generate the optimal object packing order, such as the genetic algorithm (GA) [6], generally resulting in highly computational cost and relatively low accuracy. Learning-based methods [7, 8, 9] employing DRL to resolve the BPP have gradually come up in academia, being advantageous in saving computational cost as well as increasing packing accuracy.

Some achievements have been witnessed in bin packing optimization from substantial literature available during the past few decades. The earliest research on the BPP dates back to the 1960s [10], in which a collection of objects with various weights is packed into the minimum number of boxes with the same capacity, called one-dimensional (1D)-BPP. All the objects were simply sorted in descending order of weight to increase the space utilization and reduce the usage number of boxes [11]. However, most of the objects in realistic scenarios have multi-dimensional dimensions rather than scalar weights, enabling the 1D-BPP to be unavailable in practical scenarios. The 2D-BPP and 3D-BPP are the extensions of the 1D-BPP, where the objects and boxes do not have scalar weights or capacities but 2D or 3D dimensions. Conventional packing methods of 2D-BPP and 3D-BPP for objects with regular shapes leverage meta-heuristics, including the GA, the local search (LS) algorithm [12], and the tabu search (TS) algorithm [13] to optimize the object packing order. The GA is widely applied in combinatorial optimization problems. Wang [14] presented an adaptive GA for the 2D-BPP and a hybrid GA was proposed with a heuristic algorithm [15] for the 3D-BPP in the same year. The LS algorithm selects the current solution from a neighborhood solution, until a local optimal solution is reached. Imahori *et al.* [16] proposed the LS algorithms for the rectangle packing problem, arranging given rectangles without overlap in the plane, to minimize the spatial cost. To resolve the 3D-BPP, a heuristic algorithm based on a guided LS was presented in [17]. The TS algorithm is an extension of LS, which marks the local optimal solution and avoids it in the subsequent search. Hamiez *et al.* [18] developed a reinforced TS algorithm to address a variant of the 2D-BPP, namely the 2D strip packing problem, while Viegas *et al.* [19] employed a TS and best-fit decreasing algorithms to resolve a real-world steel-cutting problem. Unfortunately, the highly computational cost and relatively low accuracy limit the popularization of the conventional packing methods in practical scenarios. Recently, learning-based methods equipped with DRL have gradually been applied to resolving the BPP. Hu *et al.* [20] developed a neural optimization solution based on deep reinforcement learning to address the transport-and-pack (TAP) problem. To deal with the 3D-BPP having limited information about the objects to be packed, an effective and easy-to-implement constrained DRL method under the actor-critic framework was proposed in [9].

Baoying Wang and Huixu Dong are with Grasp Lab of the Mechanical Engineering Department, Zhejiang University, Hangzhou, 310058, China (Corresponding author e-mail: huixudong@zju.edu.cn).

Moreover, Song *et al.* [21] added synergies between packing and unpacking into the constrained DRL method, further improving the packing accuracy. However, the learning-based methods generally focus on optimizing the placement of objects arriving successively on a conveyor belt, in the boxes, rather than the object packing order, making them difficult to deploy in the practical scenarios with batches of objects.

How well the objects can be packed into boxes is not only determined by the object packing order, but also influenced by their arrangement within the boxes. The placement strategy is applied to arranging the ordered objects in the boxes. For the 1D-BPP, the placement strategy is generally heuristic to seek for the approximate solution, such as the Next Fit (NF) algorithm [22] and the First Fit (FF) algorithm [23]. For the 2D-BPP, Jakobs [24] first introduced the Bottom Left (BL) algorithm where each object is moved as far as possible to the bottom of the box and then to the left. To reduce the empty areas in the layout generated by the BL, Hopper [25] then developed the Bottom Left with Fill (BLF) algorithm. Karabulut *et al.* [4] further extended the BLF algorithm to the 3D-BPP, proposing the Deepest Bottom Left with Fill (DBLF) algorithm. The concept of Corner Point (CP) [26] was first introduced in a branch-and-bound method and Crainic *et al.* [27] extended it to the Extreme Point (EP). The Empty Maximal Space (EMS) [28] and Maximum Contact Area (MCA) [29] were proposed to deal with the 2D-BPP and 3D-BPP, aiming to make full use of the gaps in the boxes and extremely reduce the space waste. However, these methods do not conform to the top-down operation mode. When a robot is involved in packing tasks, the target positions determined via these methods may be inaccessible due to being blocked by the objects above them, making them different to apply in automated industrial scenarios.

Motivated by the unresolved issues in the aforementioned areas, in this work, we propose a novel optimization methodology of the 2D-BPP and 3D-BPP for objects with regular shapes. We provide the statements of the regular 2D-BPP and 3D-BPP in this work, where the objects to be packed have rectangular and cuboid shapes respectively. The key to optimizing them is to generate an object packing order by DRL and then arrange the ordered objects in the boxes via a placement strategy. Firstly, to maximize the space utilization and minimize the usage number of boxes, we construct an end-to-end DRL architecture modeled by a modified Pointer Network consisting of an encoder, a decoder, and an attention module to generate the optimal object packing order. Secondly, we pack the ordered objects onto the target positions in the boxes via a placement strategy based on a height map, which represents the placement configuration of all the objects in the boxes, preventing the objects from colliding with the inside of the boxes or other objects in the boxes. Since the target position is not blocked by other objects, the object can directly reach the target position from above, conforming to the top-down operation mode. Then the reward and loss functions for the object packing order and placement strategy are defined as the indicators of the compactness, pyramid, and usage number of boxes to conduct the proposed DRL model to generate the optimal object packing order based on an on-policy actor-critic framework. Finally, a series of experiments are implemented to compare our method with conventional packing methods to demonstrate the superiority of our method in both packing accuracy and efficiency.

We **highlight** the **novelties** of our work. **Foremost**, our core contribution is resolving the optimization of the regular 2D-BPP and 3D-BPP via DRL, increasing the packing accuracy and dramatically improving the packing efficiency compared to the GA. The **first** novelty is that we employ a DRL neural network modeled by a modified Pointer Network to generate the optimal object packing order, which has hardly been attempted in the existing literature. A placement strategy based on a height map is proposed to arrange the ordered objects in the boxes, conforming to the top-down operation mode, which is attributed to the **second** novelty. In terms of the **third** contribution, we carry out various experiments to evaluate the performance of the proposed approach.

## II. METHODOLOGY

In this section, we first briefly introduce the statements of the regular 2D-BPP and 3D-BPP. Then the architecture of the DRL modeled via a modified Pointer Network is proposed to generate the optimal object packing order. Further, we introduce the placement strategy based on a height map to pack the ordered objects onto the target positions in the boxes. Finally, we design the reward and loss function of the DRL model based on an on-policy actor-critic framework.

### A. Problem Statement

We focus on the optimization of the regular 2D-BPP and 3D-BPP with known dimension information of objects and boxes, in which the shapes of the objects to be packed are rectangular and cuboid respectively. The sizes of the objects in the regular 2D-BPP and 3D-BPP are provided by the following equations

$$l_2^i \leq L_2/2, h_2^i \leq H_2/2 \quad (1)$$

$$l_3^i \leq L_3/2, w_3^i \leq W_3/2, h_3^i \leq H_3/2 \quad (2)$$

where $l_2^i$ and $h_2^i$ respectively represent the length and height of the $i^{th}$ rectangular object, while $L_2$ and $H_2$ are the length and height of the boxes in the regular 2D-BPP. $l_3^i$, $w_3^i$, and $h_3^i$ indicate the length, width, and height of the $i^{th}$ cuboid object, while $L_3$, $W_3$, and $H_3$ are the length, width, and height of the boxes in the regular 3D-BPP respectively.

Furthermore, while arranging the objects in the boxes, we consider the physical stability as a simple criterion, in which the objects are physically stable when over 50% of their bottoms are supported.

### B. Deep Reinforcement Learning Architecture

Inspired by a multi-objective optimization model [30], we employ a DRL neural network modeled via the modified Pointer Network mentioned to generate the optimal object packing order. We first introduce the input and output structure of the modified Pointer Network, which are formulated as Eq. (3), Eq. (4), and Eq. (5)

$$X = \{s^1, s^2, \ldots, s^n\} \quad (3)$$

$$s^i = (s_1^i, s_2^i, \ldots, s_M^i)^T, i = 1, 2, \ldots, n \quad (4)$$

$$Y = \{\rho_1, \rho_2, \ldots, \rho_n\} \quad (5)$$

where $X$ represents the input sequence of the neural network, and $n$ is the number of objects to be packed. $s^i$ is a $M$-dimensional vector, which represents the dimensional information of the $i^{th}$ object. Especially, in the regular 2D-BPP, $M$ is 2, i.e. each object is modeled as a two-dimensional vector, where $s_1^i$ and $s_2^i$ represent the length and height of the $i^{th}$ object respectively, thus the input sequence $X$ is modeled as a $2 \times n$ matrix. In the regular 3D-BPP, $M$ is 3, i.e. each object is modeled as a three-dimensional vector, where $s_1^i$, $s_2^i$, and $s_3^i$ represent the length, width, and height of the $i^{th}$ object respectively, thus the input sequence $X$ is modeled as a $3 \times n$ matrix. The input structure of the regular 2D-BPP and 3D-BPP are shown in Fig. 2. $Y$ is the output sequence of the neural network, representing the object packing order. For instance, the output sequence {3, 1, 5, 4, 2} indicates that the objects are packed into the boxes in the order of Object 3 — Object 1 — Object 5 — Object 4 — Object 2.

Then we introduce the architecture of the modified Pointer Network. Specifically, the neural network is composed of an encoder, a decoder, and an attention module, as shown in Fig. 3. The encoder extracts the dimensional information of the input sequence to high dimensional feature vectors. Integrating with the attention module, the decoder generates the output sequence, namely the object packing order by decoding the information contained in the high dimensional feature vectors.

**Encoder.** Considering the structure of the input sequence, we utilize the one-dimensional (1-D) convolution layer to construct the encoder. The number of the input channels of the 1-D convolution layer is equal to the dimension of the input sequence, i.e. two and three, with concerning the regular 2D-BPP and 3D-BPP respectively. The output of the encoder is a $n \times d_h$ matrix, where $n$ represents the length of the input sequence i.e. the number of the objects in a collection to be packed and $d_h$ is the number of the output channels of the 1-D convolution layer. It is noticed that all the objects in the input sequence share the parameters of the 1-D convolution layer, which means the number of the parameters in the encoder is fixed and does not increase with the length of the input sequence, making the encoder robust.

**Decoder.** Different from the encoder, the decoder requires a recurrent neural network (RNN) architecture with memory capability because we need to consider the previously selected objects when selecting the current object. The gated recurrent neural network (GRU) able to capture widely spaced dependency in the sequential data and prevent gradient explosion is employed to construct the decoder. The GRU is not directly used to obtain the object packing order, but its hidden states storing the information of previously selected objects together with the encoding states of the input sequence are used to calculate the conditional probabilities formulated as Eq. (6) and Eq. (7) to generate the object packing order. Moreover, we take the dimensional information of the currently selected object as the input of GRU as follows

$$u_j^{t+1} = v^T tanh(W_1 e_j + W_2 h_{t+1}), j \epsilon (1, 2, \ldots, n) \quad (6)$$

$$P(\rho_{t+1} = j \mid \rho_0, \rho_1, \ldots, \rho_t) =$$

$$\begin{cases} \frac{e^{u_j^{t+1}}}{\sum_{i=1}^n e^{u_i^{t+1}}}, j = 1, 2, \ldots, n, j \neq \rho_1, \ldots, \rho_t \\ 0 \times \frac{e^{u_j^{t+1}}}{\sum_{i=1}^n e^{u_i^{t+1}}}, j = \rho_1, \ldots, \rho_t \end{cases} \quad (7)$$

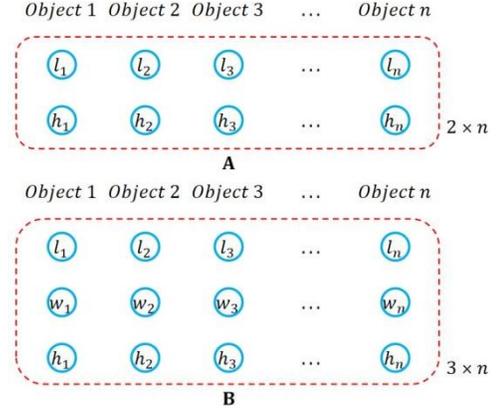

Figure. 2. Input structure of the modified Pointer Network. (A) The input $X$ in the regular 2D-BPP is a $2 \times n$ matrix. (B) The input $X$ in the regular 3D-BPP is a $3 \times n$ matrix.

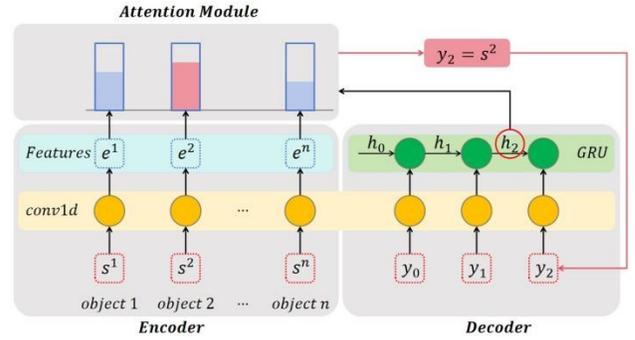

Figure. 3. The architecture of the modified Pointer Network, which consists of an encoder, a decoder, and an attention module. The attention module, integrating with the encoder and decoder, computes the conditional probabilities to generate the object packing order. As shown in the figure, the conditional probability of the object 2 is the highest, while selecting the second object to be packed, so we set $\rho_2 = 2$ and take the dimensional information of the object 2 as the input of GRU to select the third object $\rho_3$.

$$y_{t+1} = s^{\rho_{t+1}} \quad (8)$$

where $v$, $W_1$, and $W_2$ represent the learnable parameters in the attention module. $h_{t+1}$ is the hidden state output by GRU in the $t^{th}$ step. For each object $j$, $u_j^{t+1}$ is obtained via $h_{t+1}$ and its encoding state $e_j$ to calculate the conditional probability $P(\rho_{t+1} = j \mid \rho_0, \rho_1, \ldots, \rho_t)$ of selecting the object $j$ as $\rho_{t+1}$, where $\rho_0, \rho_1, \ldots, \rho_t$ are the previously selected objects. $\rho_{t+1}$ is the currently selected object, and $s^{\rho_{t+1}}$ indicates its dimensional information, which is taken as the input of GRU in the $t + 1$ step, i.e. $y_{t+1}$. In particular, we take a zero vector as the initial input of GRU, i.e. $y_0 = \vec{0}$, which means we randomly select the object $\rho_1$ from the input sequence. Furthermore, as one object can not be selected twice in a packing instance, we stipulate a masking mechanism where the conditional probabilities of the previously selected objects are multiplied by zero to prevent them from being selected twice.

**Attention module.** The attention module works with the decoder to calculate the conditional probabilities. During the training process, we randomly sample an object as $\rho_{t+1}$ according to the distribution of the conditional probabilities.

When in the testing process, we select the object with the highest conditional probability as $\rho_{t+1}$.

*C. Placement Strategy*

We need to arrange the ordered objects in the boxes via the proposed placement strategy after obtaining the optimal object packing order. Conforming to the top-down operation mode, the placement strategy based on a height map is used to pack the ordered objects onto the target positions, preventing the objects from colliding with the inside of the boxes or other objects in the boxes.

The height map represents the placement configuration of all the objects in the box. For the regular 2D-BPP, we discrete the bottom edge of the box into a one-dimensional array with length $L$ along the direction of the box length, *i.e.* X-axis, and each element in the array denotes the height of the objects in the box, generating the 2D height map $\mathbf{H}_2 \in \mathbb{Z}^L$, as illustrated in Fig. 4(A). For the regular 3D-BPP, we discrete the bottom area of the box into a two-dimensional matrix with size $L \times W$ along the direction of the box length and width, *i.e.* X- and Y-axis, and each element in the matrix denotes the height of the objects in the box, generating the 3D height map $\mathbf{H}_3 \in \mathbb{Z}^{L \times W}$ (see Fig. 4(B) and Fig. 4(C)).

When determining the target positions where objects are packed in the box, we follow the principle of minimizing the height map, which is detailed as follows. For the regular 2D-BPP, the object is packed with its bottom-left corner onto the target position, which is conjointly determined by the 2D height map, object size, and box size. We first find out all the allowable positions, which ensure no collision between the object and either the inside of the box or other objects in the box as well as keeping the object stable. The target position is the position that has a minimum $z$ value and $x$ value among all the allowable positions. We further choose the one with a minimum $x$ value as the target position when some allowable positions have equal $z$ values. Fig. 5(A) illustrates how to determine the target position in detail for the regular 2D-BPP. For the regular 3D-BPP, the object is packed with its front-bottom-left corner onto the target position, which is conjointly determined by the 3D height map, object size and, box size. Similar to the 2D-BPP, we first find out all the allowable positions and the target position is the position that has a minimum $z$ value, $y$ value, and $x$ value among all the allowable positions. We further choose the one with a minimum $y$ value and $x$ value as the target position when some allowable positions have equal $z$ values. Subsequently, we pack the bottom-left/ front-bottom-left corner of the object onto the target position and update the height map to pack the next object.

*D. Loss Function*

We optimize the solution of the regular 2D-BPP and 3D-BPP *i.e.* maximize the space utilization and minimize the usage number of boxes via a DRL neural network. To achieve this objective, we design the reward function of the proposed DRL model by integrating the compactness, pyramid, and usage number of boxes, which is formulated as

$$R = 5 * \left[\alpha \cdot \left(1 - \frac{\sum_{i=1}^{k} C_i}{k}\right) + \beta \cdot \left(1 - \frac{\sum_{i=1}^{k} P_i}{k}\right)\right] \quad (9)$$

where $k$ indicates the usage number of boxes, while $C_i$ and $P_i$ are respectively the compactness and pyramid of the $i^{th}$ box.

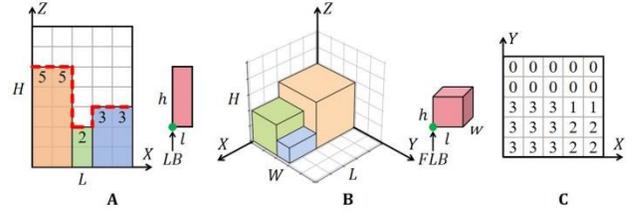

Figure. 4. The illustration of the 2D and 3D height map. (A) The left is the height map of a regular 2D-BPP instance, which is a one-dimensional array, $\mathbf{H}_2 = [5\ 5\ 2\ 3\ 3]$. The right shows the bottom-left corner of the object to be packed. (B) The left displays the placement configuration of the objects in the box of a regular 3D-BPP instance. The right shows the front-bottom-left corner of the object to be packed. (C) The height map of the regular 3D-BPP instance in (B), which is a two-dimensional matrix.

$\frac{\sum_{i=1}^{k} C_i}{k}$ and $\frac{\sum_{i=1}^{k} P_i}{k}$ represent the average compactness and average pyramid of the $k$ boxes, respectively. $\alpha$ and $\beta$ are the hyperparameters to tradeoff the compactness and pyramid in the reward function. The instructions for the compactness and pyramid are detailed as follows.

Compactness $C$ is the ratio between the total area/ volume of all objects in the box and the region defined by the maximum height in the box, as illustrated in Fig. 5(B, D). The region should be minimized to fully leverage the space in the box. Therefore, we encourage maximizing the compactness to improve the space utilization. Pyramid $P$ is the ratio between the total area/ volume of all objects in the box and the region acquired by projecting all objects to the bottom of the box, as shown in Fig. 5(C, E). The objects packed in the box should leave more continuous space available to pack the subsequent objects to enhance the space utilization. Thus, we encourage maximizing the pyramid to avoid continuous space being blocked by the objects.

We train the DRL neural network via an on-policy actor-critic framework, where two networks, namely the actor-network and critic-network are constructed (see Fig. 6). In this work, the actor-network is the modified Pointer Network introduced above, taking action *i.e.* generating the object packing order, and the critic-network consisting of four 1-D convolution layers is used to evaluate the behavior of the actor-network. The bottom 1-D convolution layer of the critic-network shares the parameters with the encoder of the actor-network, and the critic-network ultimately outputs a value that represents the expected reward the actor network can acquire concerning a given input sequence. The on-policy method means that the actor-network to train and the actor-network for interacting are the same. For each batch, we use the actor-network with the current parameters to generate the object packing order and compute the reward. Then we update the parameters of the actor-network by the policy gradient. The parameters of the critic-network are updated by reducing the difference between the real reward obtained via the actor-network and the expected reward predicted via the critic-network. The corresponding loss functions of the actor-network and critic-network are formulated as

$$L_{actor} = [R - V^\theta(X, \emptyset)] * \sum_{t=0}^{n-1} \log P(\rho_{t+1}, \theta | \rho_0, \rho_1, \ldots, \rho_t)$$

$$L_{critic} = [R - V^\theta(X, \emptyset)]^2 \quad (10)$$

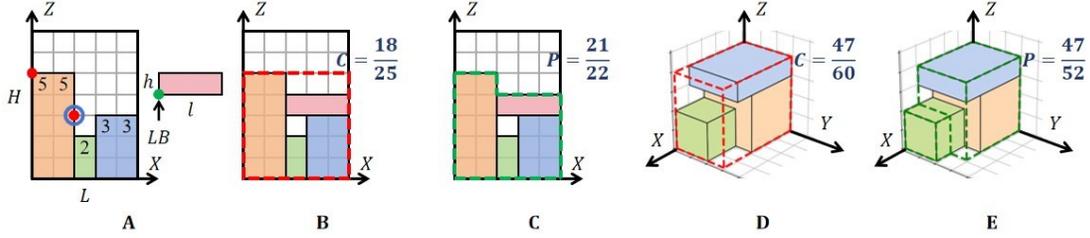

Figure. 5. The illustration of the placement strategy and the compactness and pyramid in the reward function. The pink rectangular represents the object to be packed and the green solid dot is its bottom-left corner. The red solid dot indicates the allowable positions (0, 5) and (3, 2), and the position (3, 2) with the minimum $z$ and $x$ values is chosen as the target position, highlighting with a blue hollow circle (A). The compactness $C$ is the ratio between the total area/ volume of all objects in the box and the region (red dashed line) defined by the maximum height in the box (B, D). The pyramid $P$ is the ratio between the total area/ volume of all objects in the box and the region acquired by projecting all objects to the bottom of the box (green dashed line), which is equal to the summation over the height map (C, E).

where $\theta$ and $\emptyset$ are respectively the parameters of the actor-network and critic-network, while $R$ is the reward calculated via Eq. (9). $P(\rho_{t+1}, \theta | \rho_0, \rho_1, \ldots, \rho_t)$ is the conditional probability computed by the actor-network with the parameter $\theta$ via Eq. (6) and Eq. (7). As the output of the critic-network, $V^\theta(X, \emptyset)$ represents the expected reward that the actor-network with the parameter $\theta$ can acquire concerning a given input sequence $X$.

## III. EXPERIMENTS

In this section, we carry out various experiments on the regular 2D-BPP and 3D-BPP to evaluate the performance of the proposed methods. We first describe the implementation details in the experiments, including the construction of the datasets and the configuration of the DRL neural network. Then the evaluation metrics are introduced for the optimization performances of the regular 2D-BPP and 3D-BPP. Finally, we illustrate the optimization performances obtained via our method and conventional packing methods, verifying the superiority of our method in both packing accuracy and efficiency.

### A. Implementation Details

The implementation details about the construction of the datasets in this work are introduced as follows. For the regular 2D-BPP, we set the number of objects to be packed to 40, the lengths and heights of which are randomly sampled from a uniform distribution of 1–5, respectively. Moreover, All the boxes have the same size (10, 10). For the regular 3D-BPP, we set the number of objects to be packed to 70, the lengths, widths, and heights of which are randomly sampled from a uniform distribution of 2–5, respectively. Moreover, All the boxes have the same size (10, 10, 10).

Following the aforementioned statements, we randomly generate 100,000 instances as the training set and 10,000 instances as the validation set to train the DRL neural network for the optimization of the regular 2D-BPP and 3D-BPP, respectively. We use the Adam optimizer to update the parameters of the actor-network and critic-network. The learning rate is set to $5 \times 10^{-4}$ for both actor-network and critic-network. The batch size is set to 50 while the total number of training epochs is set to 5. Thus, during the training process, the parameters of the actor-network and critic-network are updated 20,000 times in total. The hyperparameters $\alpha$ and $\beta$ in the reward function are assigned with 0.5 and 0.5 in the regular 2D-BPP and 3D-BPP, which means we treat the compactness and pyramid as equally important.

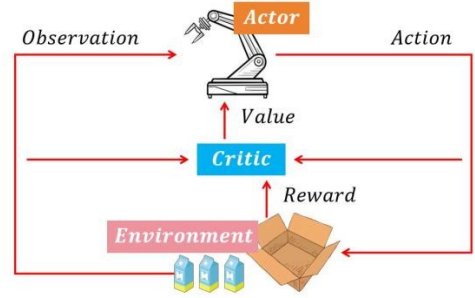

Figure. 6. The actor-critic framework to train the DRL neural network. The actor take the observation of the environment, *i.e.* the dimensional information of the objects and boxes as input, and take action *i. e.* generate the object packing order, obtaining corresponding reward. The critic evaluate the behavior of the actor, predicting the expected reward that the actor can acquire.

### B. Evaluation Metrics

To evaluate the optimization performances, we employ three indicators, including the average compactness, average pyramid, and usage number of boxes. The average compactness displays the space utilization of the boxes while the average pyramid measures the available space in the boxes to pack the subsequent objects, so we encourage maximizing the average compactness and average pyramid as well as minimizing the usage number of boxes. Meanwhile, we take the operation time to complete per instance as another indicator to have an intuitional evaluation of the packing efficiency.

### C. Comparison with Conventional Packing Methods

We compare our method with conventional packing methods which respectively generate the object packing order randomly (Random), sorting the objects in descending order of area/ volume (B-Box Seq) [31], and employing the Biased Random Key Genetic Algorithm (BRKGA) [32] to optimal the object packing order. Moreover, the placement strategy based on a height map is applied to arranging the ordered objects in the boxes.

Figure. 7 shows the optimization performances of a regular 2D-BPP instance where the same collection of rectangular objects is packed into boxes of the same size, using the conventional packing methods (Random, B-Box Seq, and BRKGA) and our method (DRL). Obviously, only our method uses the minimum number of boxes to accommodate all the objects. Table I displays the average compactness, average pyramid, usage number of boxes, and the operation time to complete per instance of the regular 2D-BPP on the average of 1,000 BPP instances, obtained via the conventional packing methods and our method. Compared to the BRKGA, our method increases the average compactness by 0.019 (0.929 vs. 0.910) and decreases the usage number of boxes by 0.009 (4.203 vs. 4.212). Moreover, our method far outperforms the BRKGA in the operation time, which saves the operation time more than 1000 times.

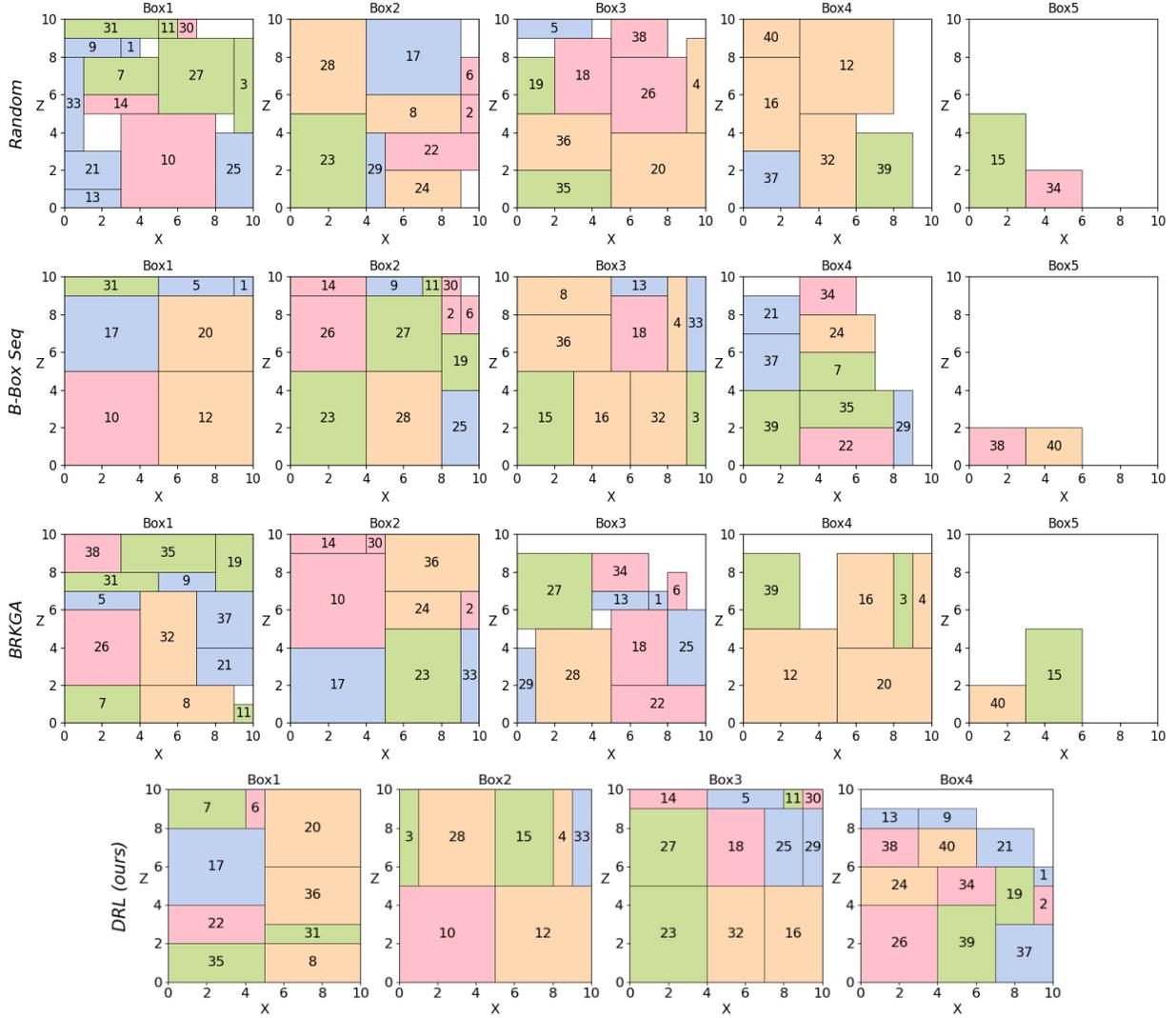

Figure. 7. The optimization performances of the regular 2D-BPP via the methods of Random, B-Box Seq, BRKGA and DRL (ours) respectively from top to bottom.

Figure. 8 shows the optimization performances of a regular 3D-BPP instance where the same collection of cuboid objects is packed into boxes of the same size, using the conventional packing methods (Random, B-Box Seq and BRKGA) and our method (DRL). Obviously, only our method uses the minimum number of boxes to accommodate all the objects. Table II displays the average compactness, average pyramid, usage number of boxes, and the operation time to complete per instance of the regular 3D-BPP on the average of 1,000 BPP instances, obtained via the conventional packing methods and our method. Compared to the BRKGA, our method increases the average compactness by 0.022 (0.791 vs. 0.769) and decreases the usage number of boxes by 0.01 (4.078 vs. 4.088). Moreover, our method far outperforms the BRKGA in the operation time, which saves the operation time more than 1000 times.

TABLE I. OPTIMIZATION PERFORMANCE OF THE REGULAR 2D-BPP ON THE AVERAGE OF 1,000 INSTANCES, OBTAINED VIA THE METHODS OF RANDOM, B-BOX SEQ, BRKGA AND DRL (OURS), WHERE $C$, $P$, AND NUM. REPRESENT THE AVERAGE COMPACTNESS, AVERAGE PYRAMID, AND USAGE NUMBER OF THE BOXES RESPECTIVELY WHILE LAT. IS THE OPERATION TIME TO COMPLETE PER INSTANCE.

| Methods | $C$ | $P$ | Num. | Lat.(ms) |
|---|---|---|---|---|
| Random | 0.832 | 0.949 | 4.646 | 0.672 |
| B-Box Seq | 0.894 | 0.964 | 4.307 | 0.927 |
| BRKGA | 0.910 | 0.990 | 4.212 | 832.715 |
| DRL | 0.929 | 0.985 | 4.203 | 0.764 |

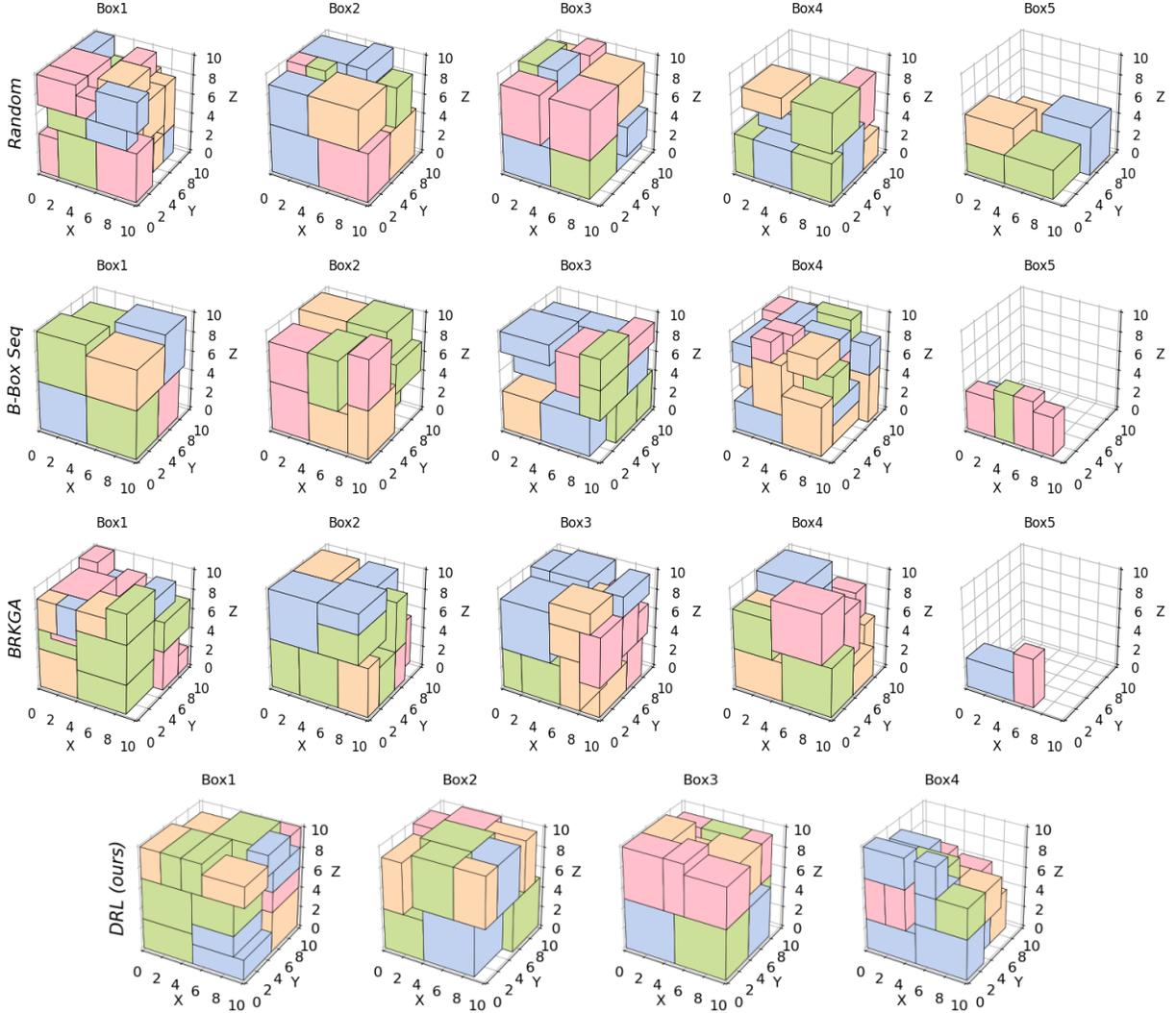

Figure. 8. The optimization performances of the regular 3D-BPP via the methods of Random, B-Box Seq, BRKGA and DRL (ours) respectively from top to bottom in the figure.

## IV. CONCLUSION

In summary, this work mainly focuses on the optimization of the Bin Packing Problem. We present a novel optimization methodology via a DRL neural network modeled by a modified Pointer Network for both 2D-BPP and 3D-BPP for regular objects. A series of experiments implemented to compare our method with conventional packing methods demonstrate the superiority of our method in both packing accuracy and efficiency.

The limitations of this work can be summarized as follows. First, we focus on the discrete BPP where the dimension of the objects and boxes are both integers. When deploying our method to realistic application scenarios, the dimension of the practical objects requires to be approximated by rounding up, which leads to the gaps in the boxes and the decline of the space utilization. Second, we limit the optimization to regular objects with rectangular or cuboid shapes, while the objects to be packed in realistic application environments are usually irregular. To deal with the irregular objects, we need to approximate the irregular objects to their minimum bounding boxes, resulting in space waste and possibly unstable placement. Given the above limitations, we assume to extend our method from discrete to continuous as well as from the regular objects to the irregular objects in future work.

TABLE II. OPTIMIZATION PERFORMANCE OF THE REGULAR 3D-BPP ON THE AVERAGE OF 1,000 INSTANCES, OBTAINED VIA THE METHODS OF RANDOM, B-BOX SEQ, BRKGA AND DRL (OURS), WHERE $C$, $P$, AND NUM. REPRESENT THE AVERAGE COMPACTNESS, AVERAGE PYRAMID, AND USAGE NUMBER OF THE BOXES RESPECTIVELY WHILE LAT. IS THE OPERATION TIME TO COMPLETE PER INSTANCE.

| Methods | $C$ | $P$ | Num. | Lat.(ms) |
|---|---|---|---|---|
| Random | 0.677 | 0.929 | 4.683 | 37.075 |
| B-Box Seq | 0.747 | 0.964 | 4.253 | 44.323 |
| BRKGA | 0.769 | 0.958 | 4.088 | 48463.754 |
| DRL | 0.791 | 0.961 | 4.078 | 39.583 |